\documentclass[a4paper]{article}

\usepackage{INTERSPEECH2022}

\title{Span Classification with Structured Information for Disfluency Detection in Spoken Utterances}
\name{Sreyan Ghosh$^{1,4}$, Sonal Kumar$^2$, Yaman Kumar Singla$^{3,4,5}$, Rajiv Ratn Shah$^4$, S. Umesh$^1$}
\address{
  $^1$Speech Lab, Department of Electrical Engineering, IIT Madras,\\
  $^2$Cisco Systems, $^3$Adobe Media Data Science Research,\\
  $^4$ IIIT-Delhi, $^5$ SUNY at Buffalo}
\email{gsreyan@gmail.com, skbrahee@gmail.com, ykumar@adobe.com, rajivratn@iiitd.ac.in, umeshs@ee.iitm.ac.in}

\begin{document}

\maketitle
\begin{abstract}
Existing approaches in disfluency detection focus on solving a token-level classification task for identifying and removing disfluencies in text. Moreover, most works focus on leveraging only contextual information captured by the linear sequences in text, thus ignoring the structured information in text which is efficiently captured by dependency trees. In this paper, building on the span classification paradigm of entity recognition, we propose a novel architecture for detecting disfluencies in transcripts from spoken utterances, incorporating both contextual information through transformers and long-distance structured information captured by dependency trees, through graph convolutional networks (GCNs). Experimental results show that our proposed model achieves state-of-the-art results on the widely used English Switchboard for disfluency detection and outperforms prior-art by a significant margin. We make all our codes publicly available on GitHub\footnote{https://github.com/Sreyan88/Disfluency-Detection-with-Span-Classification}.

\end{abstract}

\section{Introduction}
Speech is the natural way to communicate and is still the most preferred medium for communication. Unlike written text, in spoken utterances, humans often fail to pre-mediate what they are going to say, leading to interruption to speech flow. This phenomenon, called disfluency, is a para-linguistic concept that is ubiquitous in human conversations. In the past decade, with Automatic Speech Recognition (ASR) systems achieving near-human performance in transcribing speech-to-text, the use of speech as an input to modern intelligent NLU systems has democratized to an enormous extent. However, these downstream systems, trained on fluent data, can easily get misled due to the presence of disfluencies. Thus disfluency detection and removal can output clean inputs for downstream NLP tasks, like dialogue systems, question answering, and machine translation. Moreover, disfluency detection also finds applications in automatic speech scoring \cite{bamdev2022automated,kumar2019get}.

\begin{figure}[t]
  \centering
  \includegraphics[height=30pt, width=180pt]{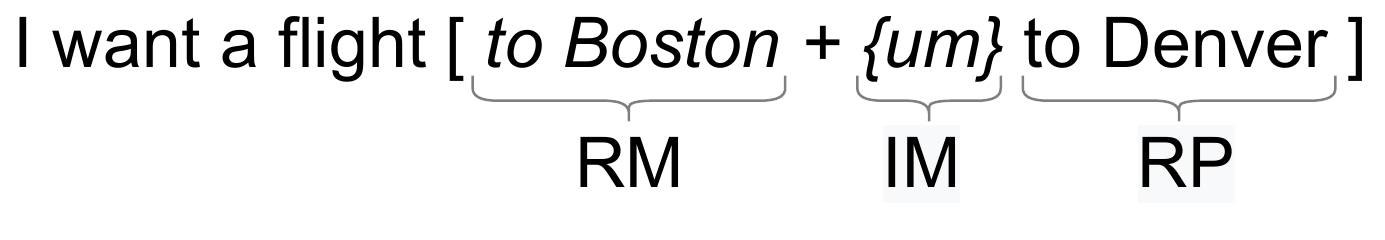}
  \caption{A sentence from the English Switchboard
corpus with disfluencies. RM=Reparandum,
IM=Interregnum, RP=Repair. The preceding RM is corrected by the following RP.
}
  \label{fig:dis_fig}
\end{figure}

Figure \ref{fig:dis_fig} shows the general structure of a disfluency, whereby it can be divided into 3 main parts, reparandum, an optional interregnum, and repair. Disfluency detection with neural architectures generally focuses on identifying and removing reparandum. Furthermore, reparandum in disfluencies can primarily be categorized into 3 types, repetitions, restarts, and repairs, as shown in Table 1. Repetition occurs when linguistic materials repeat, usually in the form of partial words, words, or short phrases. Substitution occurs when linguistic materials are replaced to clarify a concept or idea. Deletion, also known as false restart, refers to abandoned linguistics materials.

\begin{table}[ht]
    \centering
    \caption{Different types of disfluencies.}
    \begin{tabular}{|l|l|}
        \hline Type & Example \\
        \hline Repair & [ I just + I ] enjoy working \\
        \hline Repetition & {$[$ it's +$\{$ uh $\}$ it's $]$ almost like } \\
        \hline Restart & [ we would like +] let's go to the \\
        \hline Deletion & [this +$\{$ is $\}$ just +] happened yesterday.\\
        \hline Substitution & it's nothing but wood [+ $\{$up here $\}$ +]\\ & down here.\\
        \hline
    \end{tabular}
    \label{tab:dis_table}
\end{table}

In the past decade, neural architectures have shown promising results in the task of disfluency detection, where most prior works in this domain primarily report results on SwitchBoard (SWBD) corpus and solve a \emph{sequence-tagging} task. SWBD is a corpus of English telephonic conversations. The task of disfluency detection primarily focuses on detecting reparandums from segmented transcripts of individual SWBD utterances where every word in the utterance transcript is annotated as part of disfluency (or not). Most of these architectures, achieving state-of-the-art performance, solve a token-level classification task, which assigns a label to each token in the input sequence, specifying if the token is part of a disfluency or not. Though token classification is one of the most common choices for Entity Recognition (ER) and has several advantages, including not constraining the output to a single span and incorporating neighboring tag information when implemented with CRFs, we hypothesize that detecting disfluencies, especially reparandums, is different from ER where individual reparandum tokens seldom occur in isolation and make sense only when the entire span of tokens is considered together. Thus, building on the fact that models can make better semantic sense of reparandums out of spans of tokens, we propose to solve a \emph{span classification} task for disfluency detection and device models that can learn richer \emph{span} representations and not \emph{token} representations. 

The primary task of span prediction is to tag a group of one or more contiguous tokens (up to a maximum specified length) by aggregating the individual token representations into a single span representation. This kind of learning also has better boundary supervision, and since reparandums generally occur in succession to each other, we posit that this might be an added benefit to using \emph{span classification} for disfluency detection. Also, in recent years there has been a paradigm shift of ER systems for text, from token-level prediction to span-level prediction, and has seen much success \cite{fu-etal-2021-spanner}.

\begin{figure}[h]
  \centering
  \includegraphics[height=35pt, width=225pt]{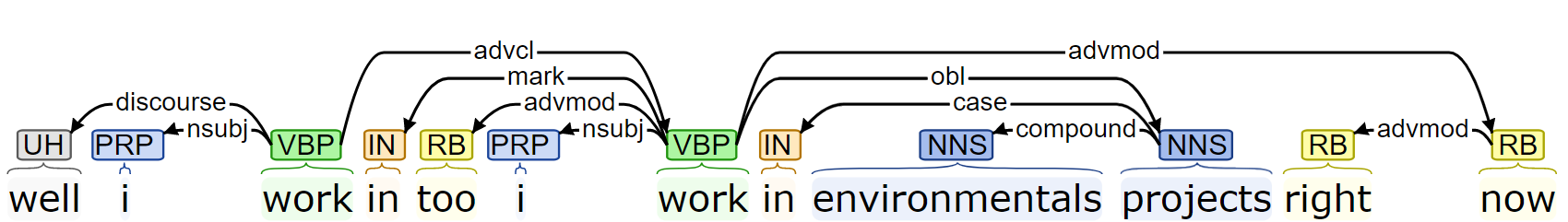}
  \caption{A sentence annotated with dependency trees}
  \vspace{-0.5em}
  \label{fig:parse_tree}
\end{figure}
LSTMs and transformer-based encoders have been two of the most common choices to encode text for the task of disfluency detection \cite{zayats2016disfluency,wang-etal-2016-neural}. However, their models suffer from certain drawbacks due to their architectural properties. Bi-directional LSTMs fail to capture long-range dependencies \cite{bengio2009learning}, and self-attention-based models can only go as far as assigning dynamically learned attention values to each token which has often been found to focus on wrong evidence for task-specific fine-tuning \cite{gupta2021my}. Thus, integrating structured information such as the interaction between nearby words or long-range word dependencies can aid the model in assigning higher importance to tokens that are syntactically related to the occurrence of another. To do this, we propose to use dependency trees to capture information along multiple dependency arcs between tokens in a sentence. For instance, as we can see in Fig. \ref{fig:parse_tree}, the span \emph{``i work in''} occurs twice in the sentence however, only the former span should be tagged as disfluent by the system. We hypothesize that the information that the primary nouns of the sentence \emph{``environmentals projects''}  only has a dependency path to the latter occurrence of the token \emph{``work''}, might help the system better integrate long-range structured dependencies and thus generate richer representations for disfluent and non-disfluent tokens.





To address the above limitations, in this paper, we propose a novel architecture based on the span classification paradigm, which leverages both contextual information using transformers and structured information obtained through dependency trees using a GCN \cite{shah2021all,zhang2018graph}. More specifically, we concatenate the graph-encoded representation of each token to the transformer output representation of that token via a gating mechanism, post which we calculate the span representation for each span of contiguous words in the input transcript. To the best of our knowledge, this is the first work that leverages structured information in text and solves a span classification task for disfluency detection in spoken utterance transcripts. To sum up, our contributions are as follows:
\begin{itemize}
\item We propose a novel architecture for the task of disfluency detection under the \emph{span-classification} paradigm of entity recognition and integrate structured information in the text to aid the model in learning better feature representations.

\item Through quantitative and qualitative analysis of the English SWBD dataset, we show that the proposed architecture and learning framework can lead to better performance of disfluency detection than solving a \emph{token-classification} task and not using structured information.

\item Our proposed learning methodology achieves state-of-the-art results for disfluency detection on the widely used English SWBD dataset.
\end{itemize}

\begin{figure}[h]
  \centering
  \includegraphics[height=310pt, width=225pt]{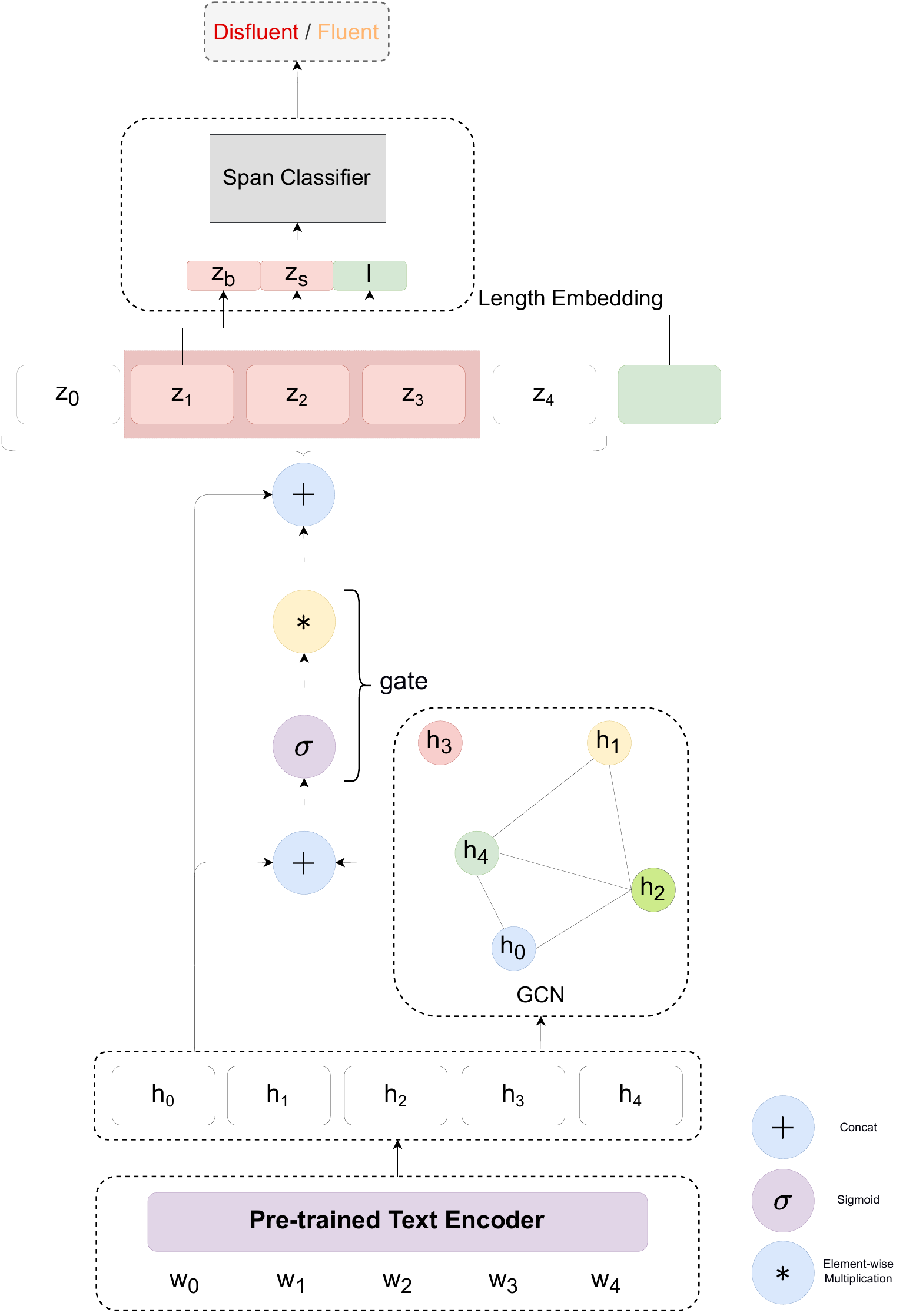}
  \caption{Proposed architecture leveraging span classification and graph-encoded structured dependency relations}
  \vspace{-1.5em}
  \label{fig:speech_production}
\end{figure}
\section{Related Work}
Disfluency detection systems can be divided into 4 primary categories. The first one makes use of noisy channel models \cite{zwarts-johnson-2011-impact,jamshid-lou-etal-2018-disfluency}, which require a Tree Adjoining Grammar (TAG) based transducer in the channel model. The second category of models leverages phrase structure, which is often related to transition-based parsing and yet requires annotated syntactic structure \cite{lou2020improving,rasooli-tetreault-2013-joint}. The third is the most common kind and frames the task as a sequence tagging task \cite{ferguson2015disfluency,hough2015recurrent}, and the last one employs end-to-end Encoder-Decoder models \cite{wang-etal-2016-neural,wang-etal-2018-semi} to detect disfluent segments automatically. In this work, we focus on improving on the third kind, which is the sequence tagging approach. Earlier work on this learning paradigm mostly focused on devising new architecture using Bi-LSTMs, and CRFs \cite{wang-etal-2016-neural}. Most recent work on this paradigm tried to alleviate the dependency on human-annotated datasets, where the authors \cite{yang-etal-2020-planning,wang2020combining} propose self-supervised learning and data augmentation approaches to learning disfluencies and has proven to close the gap with supervised training. Very recently, \cite{lee21g_interspeech} proposed to solve auxiliary sequence tagging tasks in addition to the original task of tagging disfluent tokens and serves as the current SOTA on the task of disfluency detection.

Though recent years have seen frequent paradigm shifts for the task of ER from \emph{token-level classification} to \emph{span-level classification}, this framework has been relatively under-studied. \cite{fu-etal-2021-spanner} conducts a detailed study on the complementary advantages and architectural biases provided by the latter.




\section{Proposed Methodology}
\subsection{Problem Definition}
The problem of disfluency detection can be formulated as a sequence tagging task where our primary aim is to identify tokens in spoken utterance transcripts which correspond to a reparandum. We denote the \emph{i}-th sentence with \emph{T} tokens as \emph{s\textsubscript{i}} = \{$w$\textsubscript{$t$}$|$ $t$ = 1, $\cdots$ $T$\}, and our complete dataset denoted by \{$s\textsubscript{1}$,$s\textsubscript{2}$,$s\textsubscript{3}$, $\cdots$ $s\textsubscript{$N$}$\}, where $N$ is the total number of sentences in our dataset. The corresponding label set is defined by \{$d\textsubscript{1}$,$d\textsubscript{2}$,$d\textsubscript{3}$, $\cdots$ $d$\textsubscript{$N$}\} where $d\textsubscript{i}$ is the label sequence for each sentence and is denoted by $d\textsubscript{i}$ = \{$y$\textsubscript{$t$}$|$ $y$ = 1, $\cdots$ $T$\} where $y$\textsubscript{$t$} $\in \mathcal{Y}$ and $\mathcal{Y}$ $=$ $\{I,O\}$. $I$ here stands for disfluent tokens and $O$ stands for fluent tokens.

\subsection{Proposed Architecture}

Fig. \ref{fig:speech_production} shows a pictorial representation of our proposed model architecture. As mentioned earlier, we integrate both
structured and contextual information via token representations obtained from what we call  Contextual and Structured Information layers, respectively. Post extraction and integration of these features, we pass them to a classification layer and follow a heuristic decoding strategy for tagging our sequences. Our Contextual Information Layer and decoding strategy are in line with the previous implementation in the span classification paradigm \cite{jiang-etal-2020-generalizing, ouchi-etal-2020-instance, yu-etal-2020-named, li-etal-2020-unified, xue2020coarse}.

\subsubsection{Contextual Information Layer}
Given a sentence \{$w$\textsubscript{1},$w$\textsubscript{2},$w$\textsubscript{3}, $\cdots$ $w$\textsubscript{T}\} with $T$ words, the token representation layer outputs an encoding $\mathbf{h}_{i}$ for each token as follows:

\begin{equation}
\mathbf{h}_{1}, \cdots, \mathbf{h}_{T}=\operatorname{EMB}\left(w_{1}, \cdots, w_{T}\right),
\end{equation}

where $\operatorname{EMB(.)}$ is a pre-trained contextualized text embedding model from the transformers family and $\mathbf{h}_{i}$ $\in \mathbb{R}^{d}$.


\subsubsection{Structured Information Layer}

To incorporate the long-range dependencies and structured information between the tokens in the input sentence, we propose the use of an additional graph-encoded representation $g\textsubscript{i}$ for each token using dependency parse trees. We follow a simple approach whereby we use a Graph Convolution Network (GCN) to capture information along multiple dependency arcs between words in a sentence. Thus, given the context embeddings from the Bi-LSTM, the graph-encoded representation for each token $i$ is obtained as follows:

\begin{equation}
\mathbf{g}_{1}, \cdots, \mathbf{g}_{T}=\operatorname{GCN}\left(\mathbf{h}_{1}, \cdots, \mathbf{h}_{T}\right),
\end{equation}

Post this step, we implement a gate for dynamically adjusting the contribution of features contributing to structured information in tokens, obtained from the $\operatorname{GCN}$. Following the practice in previous work, we combine the representations from both Contextual and Structured information as follows:

\begin{equation}
\mathbf{g}=\sigma\left(\mathbf{W}_{g}^{\top}[ \mathbf{R} ; \mathbf{Q}] + \mathbf{B}_{g}\right)
\end{equation}

\begin{equation}
\mathbf{gate}=\sigma\left(\mathbf{W}_{a}^{\top} \mathbf{A}+\mathbf{W}_{q}^{\top} \mathbf{Q}\right)
\end{equation}

where $\mathbf{W} \in \mathbb{R}^{2d\times d}$ is a weight matrix and $\sigma$ is the element-wise sigmoid function. The final graph-encoded representation for each token $w_t$ is then obtained by $\mathbf{g}_{t}$ $=$ $\mathbf{gate} $.$ \mathbf{g}_{t}$. Finally, for each token $w_\mathbf{i}$, the final representation fed to the span representation and prediction layers is $\mathbf{z_t}$ = [$\mathbf{h}_{t}$ $;$ $\mathbf{g}_{t}$].

\subsubsection{Span Representation Layer}

After encoding the tokens in a sentence, we enumerate through all the possible $m$ spans $J$ = \{$j$\textsubscript{1}, $\cdots$ , $j$\textsubscript{i}, $\cdots$ , $j$\textsubscript{$m$}\} upto a maximum specified length (in terms of number of tokens) for sentence $s$ = \{$w$\textsubscript{1}, $\cdots$ , $w$\textsubscript{$T$}\} and then re-assign a
label $y_i$ $\in$ $\{I,O\}$ for each span $j_i$.
For example, for the sentence ``NLP is um important'', all possible spans (or pairs of start and end indices) are \{(1, 1), (2, 2), (3, 3), (4, 4), (1, 2), (2, 3), (2,4), (1, 3), (1,4)\}, and all these spans are labelled $O$ except (3, 3) which is labelled $I$. We denote $b_i$ and $s_i$ as the start and end indices of span $j_i$ respectively. We then formulate the vectorial representation of each span as the concatenation of the representations of the starting token $\mathbf{z}_{b_i}$ $\in \mathbb{R}^{2d}$, the ending token $\mathbf{z}_{s_i}$ $\in \mathbb{R}^{2d}$, and a length embedding $\mathbf{l}_{i}$ $\in \mathbb{R}^{len}$.The length embedding is implemented as a look-up table and learnt while training the model. The final vector representation for each span fed into the span prediction layer is now $\mathbf{j}_i$ $=$ [$\mathbf{z}_{b_i}$ $;$ $\mathbf{z}_{s_i}$ $;$ $\mathbf{l}_{i}$].

\subsubsection{Span Prediction Layer}

The final span representations $\mathbf{j}_i$ is then passed through a linear transformation followed by a $\operatorname{softmax}$ operation as follows:

\begin{equation}
\mathbf{P}\left(\hat{y}_{i_{k}} \mid \mathbf{j}_{i}\right)=\frac{\operatorname{score}\left(\mathbf{j}_{i}, \mathbf{y}\right)}{\sum_{y^{\prime} \in \mathcal{Y}} \operatorname{score}\left(\mathbf{j}_{i}, \mathbf{y}^{\prime}\right)},
\end{equation}



\begin{equation}
\operatorname{score}\left(\mathbf{j}_{i}, \mathbf{y}_{k}\right)=\exp \left(\mathbf{j}_{i}^{T} \mathbf{y}_{k}\right),
\end{equation}

where $\hat{y}_{i_{k}}$ is the probability that the span $j_i$ belongs to class $k$, and $\mathbf{y}_{k}$ is a learnable representation of the class $k$.


\subsubsection{Decoding}
Since our task of disfluency detection does not have overlapping spans, we follow the heuristic decoding method from literature for non-nested entities to avoid the prediction of overlapped spans. Specifically, for overlapped spans, we keep the span with the highest prediction probability and drop the others.

\section{Experiments}

\subsection{Dataset}

We evaluate our models on the human-annotated transcriptions \cite{zayats2019disfluencies} from the English Switchboard Dataset (SWBD) \cite{godfrey1992switchboard}. SWBD is one of the most widely used datasets for evaluating disfluency detection models and frameworks. Following \cite{charniak2001edit}, we split the entire dataset into training set sw23$[\star]$.dps, development set sw4[5-9]$[\star]$.dps, and
test set sw4[0-1]$[\star]$.dps. Next, for pre-processing the transcripts before feeding them into our model, we follow the pre-processing steps mentioned by \cite{hough2015recurrent} and lower-case the text and remove all punctuation and partial words.

\subsection{Experimental Setup}

All our models are implemented using the PyTorch \cite{NEURIPS2019_9015} deep learning framework. We use Flair \cite{akbik2019flair} to implement all our \emph{token-level classification} baselines. For both our \emph{token-level classification} and \emph{span classification} models we use either of BERT\textsubscript{BASE} or ELECTRA\textsubscript{BASE} as our pre-trained contextualized token embedding model, and adopt the pre-trained checkpoints and implementation from the huggingface library \cite{wolf-etal-2020-transformers}.

We fine-tune our \emph{sequence-labelling} models with a batch size of 32 using adam optimizer with an initial learning rate of $5 \times 10^{-5}$. The dimension of our length embedding $len$ in $\mathbb{R}^{len}$ is 300 and our \emph{Structured Information Layer} has 2 layers of $\operatorname{GCN}$. 

\subsection{Baselines and Compared Methods}
For baselines, we resort to \emph{token-level classification} baselines with or without Conditional Random Field (CRF) decoders \cite{lafferty2001conditional} under the \emph{IO} tagging scheme. CRF as a decoder has been a common choice for sequence labeling tasks due to its ability to incorporate neighboring label information by considering the state transition probability of neighboring labels in \emph{token-level classification} models. Following prior-art, we choose BERT and ELECTRA from the transformers family as our contextualized text encoder for both the \emph{token-level classification} and \emph{span classification} setups. 

\subsection{Experimental Results}

Table \ref{tab:result_table} shows the evaluation results on the SWBD test compared to prior-art on disfluency detection. Consistent with prior-art, we show the \emph{ F\textsubscript{1}} scores for all our approaches. As we clearly see, our best-proposed model (Span Classification w GCN) outperforms state-of-the-art (SOTA) \cite{lee21g_interspeech} by 3.8\%  and our baselines by a significant margin. Though our \emph{span classification} models alone report significant gains, integrating structured information with GCNs imporves performance over it. Here, we want to re-iterate the fact that including structured information for disfluency detection is a logical step and might steer further research in this direction. Additionally, in the next section, we also make an effort to analyze the benefits of adding structured information to our \emph{span classifier} model.




\begin{table}[ht]
    \centering
    \caption{Evaluation results of our proposed model compared to the baselines and prior-art on the Switchboard test set. The best scores are denoted in bold.}
        \begin{tabular}{lccc}
        \hline Model & $\mathrm{P}$ & $\mathrm{R}$ & $\mathrm{F\textsubscript{1}}$\\
        \hline \emph{Prior-art}  \\
        \hline Semi-CRF \cite{ferguson2015disfluency} & $90.0$ & $81.2$ & $85.4$ \\
        Bi-LSTM \cite{zayats2016disfluency} & $91.6$ & $80.3$ & $85.9$ \\
        Attention-based \cite{wang-etal-2016-neural} & $91.6$ & $82.3$ & $86.7$ \\
        Transition-based \cite{wang2017transition} & $91.1$ & $84.1$ & $87.5$ \\
        Self-supervised \cite{wang2020combining} & $93.4$ & $87.3$ & $90.2$ \\
        Self-trained \cite{lou2020improving} & $87.5$ & $93.8$ & $90.6$ \\
        EGBC \cite{bach2019noisy} & $95.7$ & $88.3$ & $91.8$ \\
        BERT fine-tune \cite{bach2019noisy} & $94.7$ & $89.8$ & $92.2$ \\
        BERT-CRF-Aux \cite{lee21g_interspeech} & $94.6$ & $91.2$ & $92.9$ \\
        ELECTRA-CRF-Aux \cite{lee21g_interspeech} & $94.8$ & $91.6$ & $93.1$\\
        \hline \emph{Our Experiments} & & & \\
        \hline
        Our Baselines & & & \\
        \hline
         Token Classification BERT & $91.8$ & $84.9$ & $88.2$ \\
         Token Classification ELECTRA & $90.8$ & $88.3$ & $89.5$ \\
         Token Classification BERT-CRF & $93.4$ & $82.6$ & $87.7$ \\
         Token Classification ELECTRA-CRF & $92.5$ & $87.2$ & $89.8$ \\
         \hline
         Span Classification BERT& $95.1$ & $93$ & $94.1$ \\
         Span Classification ELECTRA& $90.1$ & $\mathbf{94}$ & $92.3$ \\
         Span Classification BERT-GCN & $\mathbf{95.2}$ & $93.2$ & $\mathbf{94.2}$ \\
         Span Classification ELECTRA-GCN & $91.7$ & $94$ & $92.9$ \\
        \hline
        \end{tabular}
    \label{tab:result_table}
\end{table}

\subsection{Qualitative Results Analysis}

In this section, we first analyze with some example instances where \emph{span classification} does better than \emph{token-level classification}. Next, we also study the benefit of integrating graph-encoded structured information in our \emph{span classification} model. 

\textbf{Span Classification vs. Token-level Classification}-Highlighted words represent the prediction made by our \emph{span classifier} and ground truth annotations, while the underlined words represent the predictions by our \emph{token-level classifier} model.
\begin{enumerate}
    \item i work part time at night \hl{and he works} and my husband works full time days
    \item so it was an age where \hl{it was we thought} it would be good for them to have the discipline that goes \underline{\hl{with}} with having a pet
\end{enumerate}

As we clearly see in both the examples, both spans are not obvious disfluencies other than the repetition in eg. 2 which was predicted right by the \emph{token-level classifier}. We hypothesize that longer \emph{restarts} and \emph{repairs} like the ones in the examples are better captured by our \emph{span classifier}, where the group of words with proper boundary supervision helps our \emph{span classifier} better capture semantics, than our  \emph{token-level classifier} which classifies tokens in isolation.

\textbf{GCN vs w/o GCN}-Highlighted words represent the prediction made by our \emph{span classifier} with the GCN module and ground truth annotations, while the underlined words represent the predictions by our \emph{span classifier} trained without it.
\begin{enumerate}
    \item \underline{\hl{in in}} in the winter \underline{\hl{you} typically \hl{it's}} it's probably too cold to go out and do things like tennis
    \begin{figure}[ht]
      \centering
      \includegraphics[height=35pt, width=225pt]{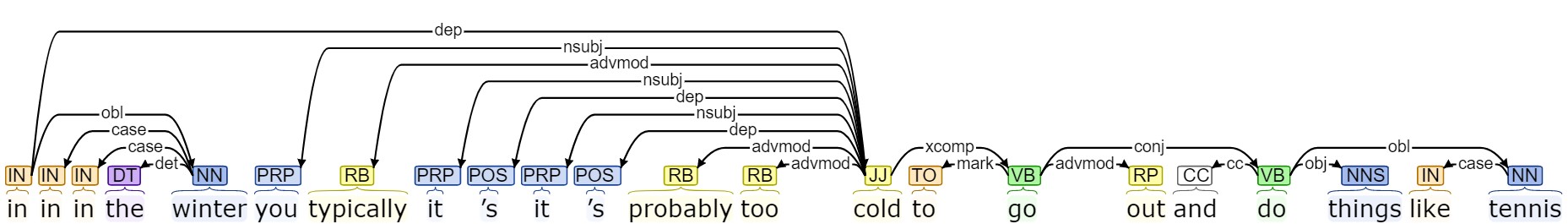}
    \vspace{-1.0em}
      \label{fig:sent3}
    \end{figure}
    \item oh even where you do have \underline{the inspections} you know the inspection is once a year
    \begin{figure}[ht]
      \centering
      \includegraphics[height=35pt, width=225pt]{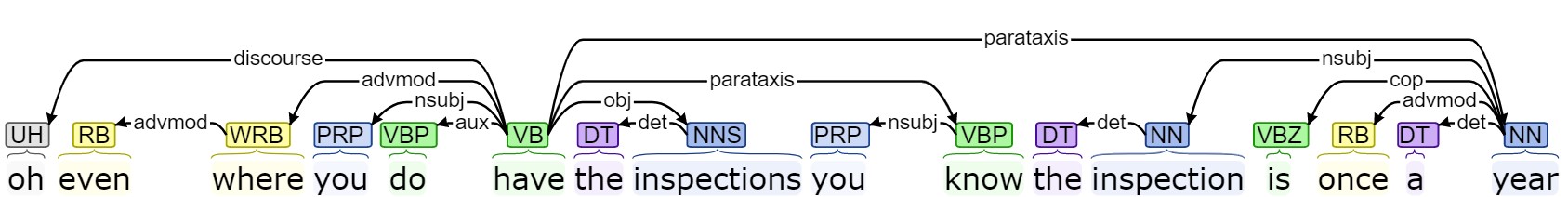}
    \vspace{-1.0em}
      \label{fig:sent4}
    \end{figure}
\end{enumerate}

In the first example above, the word ``typically'' is a difficult fluent word to capture since it occurs between 2 disfluent words. However, the GCN guides our \emph{span classifier} through the nominal subject dependency arc, which, when combined with contextual representation, helps our model understand that ``typically'' is semantically related to ``cold''. Through the second example, we also hypothesize that our \emph{span classifier w/o GCN} model might be biased to finding repeated nearby words and classify spans wrongly with any former span consisting of the repeated word. In this case, structured dependency of the former ``inspections'' with other nearby words helps the \emph{span classifier w GCN} model classify it as fluent.

\section{Conclusions}

In this paper, we propose a novel model architecture for the task of disfluency detection in spoken utterance transcripts. Our proposed model achieves SOTA on the widely used English SWBD for disfluency detection. As part of future work, we would like to investigate how to better integrate structured and contextual information for better disfluency detection.



\bibliographystyle{IEEEtran}

\bibliography{template}


\end{document}